%% file: emnlp2023.tex
\pdfoutput=1

\documentclass[11pt]{article}

\usepackage[review]{EMNLP2023}

\usepackage{times}
\usepackage{latexsym}
\usepackage{amsmath}
\usepackage{cleveref}
\usepackage{booktabs}
\usepackage{amssymb}
\usepackage{graphicx} 
\usepackage{multicol}
\usepackage{multirow}
\usepackage{booktabs}
\usepackage{diagbox}
\usepackage[T1]{fontenc}

\usepackage[utf8]{inputenc}
\usepackage[authormarkup=none,final]{changes}
\usepackage[]{todonotes}

\usepackage{microtype}

\usepackage{inconsolata}
\usepackage{devanagari}
\usepackage{colortbl}

\newcommand{\mystrut}[1]{\rule[-#1]{0pt}{#1}}

\definecolor{subjectbg}{HTML}{EA6B66}
\definecolor{verbbg}{HTML}{67AB9F}
\definecolor{objectbg}{HTML}{B5739D}

\newcommand{\equalcolorbox}[3]{
   \colorbox{#1}{\mystrut{0.08cm} #2}
}

\title{A Three-Pronged Approach to Cross-Lingual Adaptation with Multilingual LLMs}

\author{
  Vaibhav Singh$^{\spadesuit\xi}$,
  Amrith Krishna$^{\diamondsuit\lambda}$,
  Karthika N J$^{\spadesuit\xi}$,
  Ganesh Ramakrishnan$^{\spadesuit\xi}$ \\
  $^{\spadesuit}$Indian Institute of Technology Bombay,
  $^{\diamondsuit}$SML\\
    $^{\xi}$\texttt{\{singhvaibhav, karthika, ganesh\}@cse.iitb.ac.in} \\
  $^{\lambda}$\texttt{krishnamrith12@gmail.com}
}

\begin{document}
\maketitle
\bibliographystyle{acl_natbib}
\input{abstract}
\input{introduction}
\input{method}
\input{experiments}
\input{results}

\input{conclusion}
\input{limitations}
\bibliography{custom}

\input{appendix}

\end{document}

%% file: abstract.tex
\begin{abstract}
Low-resource languages, by its very definition, tend to be under represented in the pre-training corpora of Large Language Models. In this work, we investigate three low-resource cross-lingual approaches that enable an LLM adapt to tasks in previously unseen languages. \texttt{Llama-2} is an LLM where Indic languages, among many other language families, contribute to less than $0.005\%$ of the total $2$ trillion token pre-training corpora. In this work, we experiment with  the English-dominated \texttt{Llama-2} for cross-lingual transfer to three Indic languages, Bengali, Hindi, and Tamil as target languages. We study three approaches for cross-lingual transfer, under ICL and fine-tuning. One, we find that adding additional supervisory signals via a dominant language in the LLM, leads to improvements, both under in-context learning and fine-tuning. Two, adapting the target languages to word reordering may be beneficial under ICL, but its impact diminishes with fine tuning. Finally,  continued pre-training in one low-resource language can improve model performance for other related low-resource languages.

\end{abstract}

%% file: introduction.tex
\section{Introduction}
\begin{figure}[ht]
    \centering
    \includegraphics[scale=0.23]{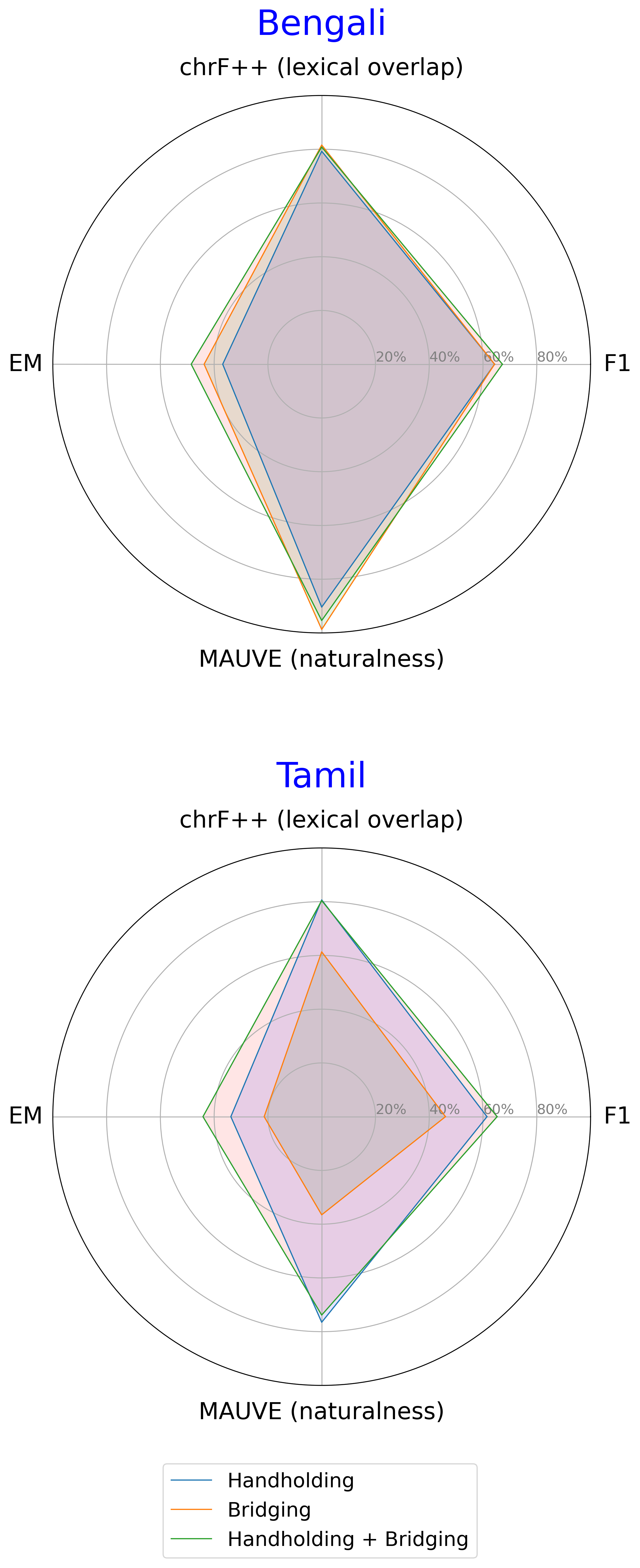}
    \caption{Improved natural language understanding (NLU) and generation (NLG) of \texttt{Llama-2-7b} in Bengali and Tamil through continued pre-training in Hindi \textit{(Bridging)} and leveraging English for cross-lingual transfer \textit{(Handholding)}.}
    \label{fig:teaser}
\end{figure}

\begin{figure*}[ht]
    \centering
    \includegraphics[width=\textwidth]{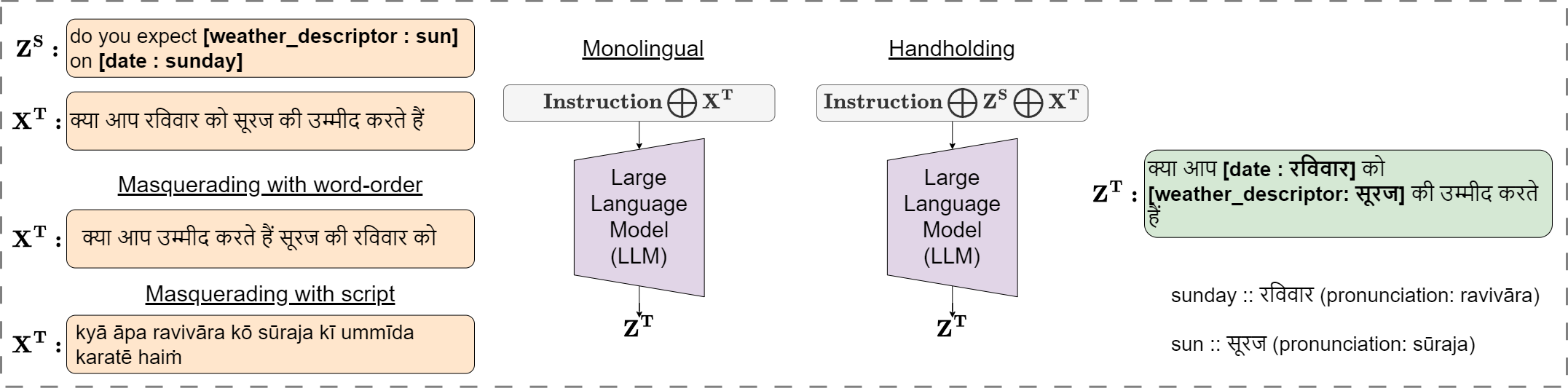}
    \caption{Task of slot filling, using the cross-lingual transfer objective from English to Hindi, using an LLM. In this example, the word `sun' translates to `sūraja' in Hindi and `sunday' translates to `ravivāra'. Thus, in the output. the LLM assigns the label $\underline{\textit{weather\_descriptor}}$ to the word `sun' in Hindi, and the label $\underline{\textit{date}}$ to `sunday' in Hindi. Refer to \Cref{train_prompt1} and \Cref{train_prompt2} for details on the prompt.}
    \label{fig:overview}
\end{figure*}

Large language models \citep[LLM;][]{brown2020language, touvron2023llama, chowdhery2022palm, gemmateam2024gemma} are known to generalise well across several tasks, including in few shot and zero-shot setups. However, there is limited evidence that shows the ability of these models to generalise to tasks in new languages out of the box, especially to those with which the model has limited exposure to.  In this work, we investigate how effectively we can leverage the LLMs for cross lingual transfer, especially for adapting it to low-resource languages.

LLMs typically require tens of billions, if not trillions, of tokens for its pre-training. Now, that is a challenge for majority of the languages in the world. More than $80\%$ of languages in the world are `left behind' \citep{joshi-etal-2020-state}, and barely have enough digitised data that matches the requirements for pre-training an LLM from scratch. For instance, the most populous country in the world, India, speaks more than $400$ languages\footnote{\url{https://en.wikipedia.org/wiki/Languages_of_India}}, with $22$ of them recognised as scheduled languages by the Government of India. However, none of these languages contribute to more than $0.005\%$ of the pre-training data of an open-source LLM like \texttt{Llama-2} \citep{touvron2023llama}. In fact, more than 95\% of these languages lack enough digital resources to incorporate them into an LLM. These resource-poor languages tend to get poorer in representation with  the progress in the field \citep{joshi-etal-2020-state, ojo2024good}.

Some of the recent works, explore various techniques to adapt an LLM to new languages, especially with limited target language resources \cite{rathore-etal-2023-zgul}. \citet{tanwar2023multilingual} exploit cross-lingual transfer to improve in-context learning (ICL) for binary sequence classification tasks in low-resource languages by utilizing in-context exemplars from a high-resource language semantically similar to the input in the target language. \citet{husain2024romansetu} employ continual pre-training on \texttt{Llama-2} with romanized pre-training corpora of non-roman script languages, to exploit cross-lingual transfer using the script of English. \citet{awasthi2023bootstrapping} use $540$b PaLM \citep{chowdhery2022palm} to generate training data in low-resource languages using labelled instances in English. \citet{razumovskaia2024analyzing} provide analyses of multilingual capabilities of LLMs on NLU tasks under the settings of in-context learning (ICL), supervised fine-tuning (SFT), and supervised instruction-tuning (SIT). 

Our investigation primarily involves the following three questions, centered around information extraction (IE) tasks in a low-resource language using an instruction-tuned LLM. \textit{Q1. Handholding:} For an IE task in a low-resource target language, would providing a parallel, annotated sentence in the predominant language of the LLM, help to exploit cross-lingual transfer, resulting in improved performance for the target language.  By predominant language, we imply the language that forms the majority of the pre-training corpora. \textit{Q2. Masquerading:} Would adapting the target language to resemble the predominant language enable in cross-lingual transfer, benefiting the target language. Finally, \textit{Q3. Bridging:} Whether model adaptation in one of the low-resource languages can benefit other related low-resource languages. More clarity on these questions, is presented in \Cref{sec:prelim}.



We focus on three Indic languages, namely, Bengali, Hindi, and Tamil. These languages are culturally diverse within the Indic context, with Bengali and Hindi belonging to the Indo-Aryan family and Tamil to the Dravidian family. To evaluate our hypotheses \textit{Q1, Q2, and Q3}, we focus on two information extraction tasks: slot filling and named entity recognition (NER). Further, we use a $7$ billion parameter English-centric LLM \texttt{Llama-2} as our base LLM, unless otherwise stated. The slot filling and named entity recognition tasks possess label-set size of $55$ and $3$, respectively. 
Additionally, none of Bengali, Hindi, and Tamil contribute to more than $0.005\%$ of the pre-training corpora of \texttt{Llama-2}. Moreover, English is the predominant language, contributing to roughly $90\%$ of the pre-training corpora.

In our experiments, we simlulate a low-resource scenario where we do not expect the target language to have more than roughly $10,000$ instances. In \textit{Bridging}, when \texttt{Llama-2} is adapted with Hindi through continued pre-training, we use more than 10,000 sentences in Hindi. However, in this case, Hindi is referred to as the bridge language.  The evaluation is solely performed on Bengali and Tamil, both of  which  satisfy aforementioned criteria for the low-resource setting. Our investigation includes exploiting few-shot in-context learning (ICL) ability of \texttt{Llama-2} as well as model adaptation with parameter-efficient supervised fine-tuning (PEFT). To evaluate \texttt{Llama-2}, or any auto-regressive LLM in general, we frame the tasks of slot filling and named entity recognition as text-to-text generation tasks. \Cref{fig:overview} showcases slot filling as a text-to-text generation task. 

Extensive experiments on \texttt{Llama-2} show that \textit{Handholding} improves NLU and NLG in Bengali, Hindi and Tamil by exploiting cross-lingual transfer from English, under both few-shot ICL and PEFT. Further, \textit{Bridging} with Hindi, improves monolingual task performance in related languages of Bengali and Tamil under PEFT. Ultimately, \textit{Handholding + Bridging} turns out the most beneficial combination, yielding best task performance for both low-resource languages of Bengali and Tamil. A quantitative overview has been presented in \Cref{fig:teaser}.

Our major contributions can be summarized as follows:
\label{sec:contri}
\begin{itemize}
    \item We demonstrate that the predominant language of an LLM can be leveraged to aid low-resource languages. Specifically, leveraging English via \textit{Handholding}, improves the overall performance of \texttt{Llama-2} for information extraction tasks in Hindi, Bengali, and Tamil under both few-shot in-context learning (ICL) and parameter-efficient fine-tuning (PEFT).
    \item Improved natural language understanding and generation in Bengali and Tamil, as shown by our experiments with \texttt{Llama-2} adapted with Hindi \textit{(Bridging)}, demonstrates that adapting a model in one low-resource language can benefit other related languages.
    \item Modifying target language  via \textit{(Masquerading)} to resemble the predominant language, English, gives superficial benefits in few-shot ICL and diminishes further in PEFT. 
\end{itemize}

%% file: method.tex
\section{Preliminaries}
\label{sec:prelim}
\subsection{Task Definition}
\label{sec:formula}
Given a finite label-set \(\mathcal{L}\), let \(\mathbf{X}^S = (X_1^S, X_2^S, \ldots, X_n^S)\) denote a sentence in source language and \(\mathbf{A}^S = (A_1^S, A_2^S, \ldots, A_n^S)\) represent the corresponding word-level label sequence, where \(A_i^S \in \mathcal{L} \cup \{\phi\}\) and \(\phi\) indicates the absence of a label. A labelled source sequence is given by \(\mathbf{Z}^S = ((X_1^S, A_1^S), (X_2^S, A_2^S), \ldots, (X_n^S, A_n^S))\). In \textit{Handholding}, our goal is to transfer these annotations to a parallel, unannotated sentence in target language \(\mathbf{X}^T = (X_1^T, X_2^T, \ldots, X_m^T)\), producing an labelled target sentence \(\mathbf{Z}^T\). \Cref{fig:overview} demonstrates the defined text-to-text cross-lingual setup. Formally,
{
\begin{equation}
    \nonumber
    \mathbf{Z}^T = \arg\max_{\mathbf{Y}}  P_{\text{LLM}}(\mathbf{Y} \mid \mathbf{Z}^S, \mathbf{X}^T)
\end{equation}
}
where \(\mathbf{Y} = ((Y_1, B_1), (Y_2, B_2), \ldots, (Y_m, B_m))\) is a potential annotated target sentence, with \(Y_i\) being elements of \(\mathbf{X}^T\) and \(B_i\) being elements of \(\mathcal{L} \cup \{\phi\}\). In our context, the conditional probability can be decomposed following the auto-regressive nature of LLM generation:
{
\begin{equation}
\nonumber
\begin{aligned}
    P_{\text{LLM}}(\mathbf{Y} \mid \mathbf{Z}^S, \mathbf{X}^T) &= \\
    &\hspace{-4em} \prod_{i} P((Y_i, B_i) \mid (Y_j, B_j)_{<i}, \mathbf{Z}^S, \mathbf{X}^T)
\end{aligned}
\end{equation}
}

In a similar manner, as shown in \Cref{fig:overview}, a monolingual objective with no \textit{Handholding}, can be formulated in the following manner:
{
\begin{equation}
    \nonumber
    \mathbf{Z}^T = \arg\max_{\mathbf{Y}}  P_{\text{LLM}}(\mathbf{Y} \mid \mathbf{X}^T)
\end{equation}
}
{
\begin{equation}
\nonumber
\begin{aligned}
    P_{\text{LLM}}(\mathbf{Y} \mid \mathbf{X}^T) =\prod_{i} P((Y_i, B_i) \mid (Y_j, B_j)_{<i}, \mathbf{X}^T)
\end{aligned}
\end{equation}
}
\subsection{Handholding, Masquerading, and Bridging}

\paragraph{Predominant Language as a Point of Supervision: }
\label{sec:compare}
In our work, with \texttt{Llama-2}, English is the predominant language with $89.70\%$ presence in the pre-training corpora of \texttt{Llama-2}. On the contrary, low-resource languages like Bengali, Hindi, and Tamil, cover less than $0.005\%$, and can be regarded as `unseen' when compared to English. To leverage the understanding of \texttt{Llama-2} in English for an IE task in a low-resource `target' language, we include annotated parallel sentence in English as a part of the task-specific prompt to the LLM. As shown in \Cref{fig:overview}, referred to as \textit{Handholding}, we utilize annotated English sentence $(\mathbf{Z^S})$ to facilitate cross-lingual transfer to the target language.

\paragraph{Adaptation of Target Language: }
To further aid cross-lingual transfer, we look at ways in which the target language can resemble English. First, we look at word order. Word order refers to the arrangement of words in a sentence. Word order is one of the syntactic features that varies across languages. English follows subject-verb-object order. On the contrary, Indic languages largely follow subject-object-verb word order where the verb appears at the tail part of a sentence. Second, we look at the script of English, to aid cross-lingual transfer. As English follows the Latin script, we employ transliteration schemes to transform the sentence in the target language to Latin. We refer to this adaptation of the target to resemble English as \textit{Masquerading}. \Cref{fig:overview} gives an overview of target sentence $(\mathbf{X^T})$ \textit{masqueraded} to resemble English.

\paragraph{Related Language as a Bridge: } Continual pre-training \citep{cui2024efficient, gupta2023continual}, vocabulary extension \citep{zhao2024llama}, instruction-tuning\citep{gala2024airavata, li2023bactrianx, husain2024romansetu} are some of the ways to increase representation of language(s) into an LLM. As Hindi is one of the most represented languages in India, we investigate the effect of adapting an LLM in Hindi through continual pre-training, on related low-resource languages of Bengali and Tamil. We refer to this as \textit{Bridging}. Hindi in this scenario, becomes the bridge language, while Bengali and Tamil become the target languages for evaulation.


%% file: experiments.tex
\section{Experiments}
\subsection{Datasets}
\label{sec:datasets}

\paragraph{Slot Filling: }We use Amazon Massive \citep{fitzgerald2022massive}. The dataset includes slot annotated virtual assistant utterances parallel across $51$ languages. We choose sentences from [\textit{utt}] and [\textit{annot\_utt}] fields of the dataset to represent unannotated sequence $\mathbf{X}$ and ground-truth annotated sequence $\mathbf{Z}$ respectively for cross-lingual transfer among languages: English, Bengali, Hindi, and Tamil. This dataset includes $55$ label types, including \texttt{place\_name, business\_name, music\_genre}, among others. Refer to \Cref{table:slot-types} for all label types and \Cref{table:dataset_split} for the train-test split.

\paragraph{Named Entity Recognition: }We work with with AI4Bharat Naamapadam \citep{mhaske-etal-2023-naamapadam}, the largest publicly available NER dataset for 11 Indic languages, sampled and annotated from Samanantar \citep{ramesh-etal-2022-samanantar}. For the languages in focus, Bengali, Hindi, and Tamil, Naamapadam has $961.7\text{k}$, $985.8\text{k}$, and $497.9\text{k}$ instances in their train split, respectively. We sample $16\text{k}$ instances for each of the languages. Due to the absence of ground-truth annotated parallel sequences in English for each of Hindi, Bengali, and Tamil, we leverage the same strategy as \citep{mhaske-etal-2023-naamapadam} and pick the corresponding set of English sentences from Samanantar and annotate them using a \texttt{bert-base} token-classification reference model. List of all label types and train-test split can be found in \Cref{table:slot-types} and \Cref{table:dataset_split}, respectively.

\subsection{Implementation Details}
\label{sec:imple_details}

To evaluate all the hypotheses presented in \Cref{sec:prelim}, we use English-centric \texttt{Llama-2-7b} \cite{touvron2023llama}. By `English-centric', we mean to point that English is the predominant language of the LLM. Particularly, we use \texttt{Llama-2-7b-chat}, the instruction-tuned variant of pre-trained base \texttt{Llama-2-7b}. The need for the instruction-tuned variant is mainly attributed to the nature of a prompt-based generation task where we expect an LLM to be prompted with an instruction followed by an input instance.

\begin{figure}[h]
    \centering
    \includegraphics[scale=0.27]{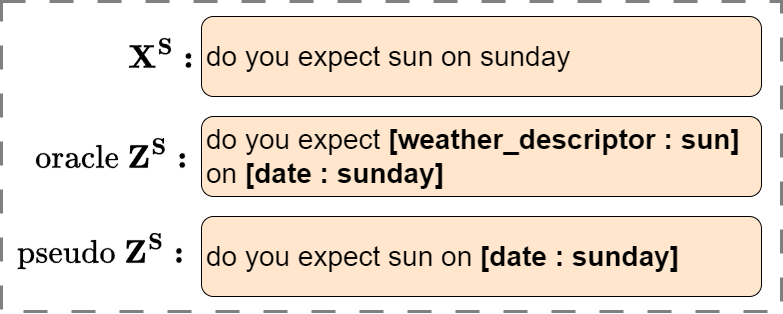}
    \caption{Here, $\text{oracle }\mathbf{Z^S}$ refers to the ground-truth annotation of $\mathbf{X^S}$. $\text{pseudo }\mathbf{Z^S}$ is obtained after passing $\mathbf{X^S}$ through an \texttt{xlm-roberta-base} token classification model.}
    \label{fig:oraclepseudo}
\end{figure}

For \textit{Handholding}, we use English as the labelled point of supervision to enable cross-lingual transfer. Further, we do not use ground-truth English labels during task-specific model inference; instead, we label the English sentence using a token classification model before the cross-lingual transfer step. We refer to these predicted labels for English as \textit{pseudo} labels and the ground-truth labels for English as \textit{oracle} labels. For slot filling, we use $84.05$ F1 score \texttt{xlm-roberta-base}\footnote{\url{https://huggingface.co/cartesinus/xlm-r-base-amazon-massive-slot}} token classification model proposed in \citep{kubis2023back}. Whereas, for named entity recognition, we use $91.3$ F1 score \texttt{bert-base}\footnote{\url{https://huggingface.co/dslim/bert-base-NER}} token classifier, as discussed in \Cref{sec:datasets}. \Cref{fig:oraclepseudo} shows the difference between an \textit{oracle} and \textit{pseudo} labelled sentence in English for the task of slot filling.

\begin{figure}[h]
    \centering
    \includegraphics[scale=0.24]{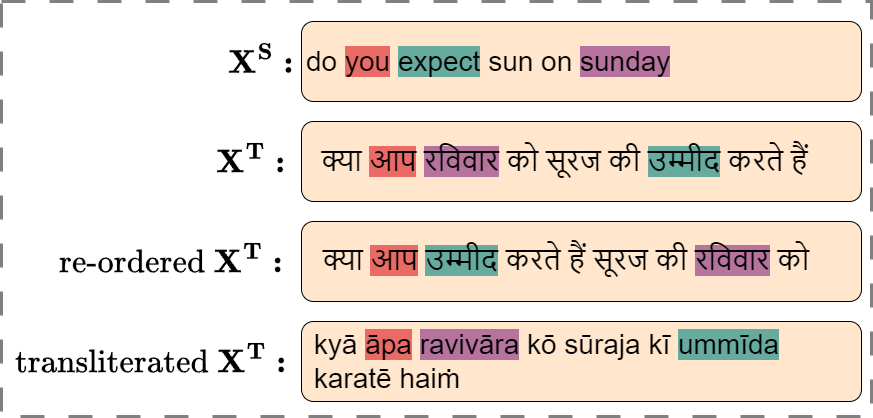}
    \caption{English follows \equalcolorbox{subjectbg}{subject}{}\equalcolorbox{verbbg}{verb}{}\equalcolorbox{objectbg}{object}{} word order in contrast to Hindi. Hindi follows the word order of \equalcolorbox{subjectbg}{subject}{}\equalcolorbox{objectbg}{object}{}\equalcolorbox{verbbg}{verb}. As shown, $\mathbf{X^T}$ is presented in SOV order and $\text{re-ordered }\mathbf{X^T}$ is presented in SVO order. $\text{transliterated }\mathbf{X^T}$ is $\mathbf{X^T}$ in Latin script using \texttt{ISO 15919:2001}. Here, only the script of $\mathbf{X^T}$ is changed, keeping the word order of Hindi.}
    \label{fig:target}
\end{figure}

In \textit{Masquerading} with word order, we use \texttt{GIZA++} \citep{och-ney-2003-systematic}, a word alignment model based on the statistical models by IBM \citep{brown-etal-1993-mathematics} and pre-trained LM-based \texttt{SimAlign} \citep{sabet2021simalign} to generate word re-ordered target sentences. Specifically, we use \texttt{SimAlign} for Hindi and \texttt{GIZA++} for Bengali and Tamil based on qualitative assessment. In the latter setting of \textit{Masquerading}, we follow \texttt{ISO 15919:2001} to transliterate the sentences in Bengali, Hindi, and Tamil to Latin script. Refer \Cref{fig:target} for an example of adapting Hindi to resemble English.

For \textit{Bridging}, we utilize \texttt{Airavata-7b} \citep{gala2024airavata}, a continually pre-trained and instruction-tuned version of pre-trained base \texttt{Llama-2-7b} model in code-mixed Hindi and English. To ensure that the effect of \textit{Bridging} in Hindi on Bengali and Tamil can be solely attributed to the increased representation of Hindi, we highlight the key differences between \texttt{Llama-2-7b-chat} and \texttt{Airavata-7b}. 

According to \citet{touvron2023llama}, \texttt{Llama-2-7b-chat} builds on \texttt{Llama-2-7b} base pre-trained model through supervised fine-tuning with publicly available SFT datasets \citep{chung2022scaling} and $27,540$ high-quality in-house vendor-based SFT annotations followed by reinforcement learning through human feedback (RLHF) \citep{ouyang2022training} with over $1$ million human annotated instances. Whereas, to train \texttt{Airavata-7b}, \citet{gala2024airavata} employ LoRA fine-tuning on a continually pre-trained \texttt{Llama-2-7b} with publicly available English SFT datasets, with their translations in Hindi, amounting to a total of $385\text{K}$ SFT instances. 

We note two observations: (1) the utilized SFT datasets do not cover either of the two datasets used in our evaluation, eliminating any case of labelled data leakage and (2) the quality of the SFT instances used for training \texttt{Airavata-7b} does not match that of \texttt{Llama-2-7b-chat}, mainly due to absence of high quality in-house annotations and the Hindi subset being translations of publicly available English SFT instances, which generally possess insufficient diversity and insufficient quality \citep{touvron2023llama}. Hereafter, we refer to \texttt{Llama-2-7b-chat} and \texttt{Airavata-7b}, simply as $\texttt{Llama}_{\texttt{chat}}$ and $\texttt{Airavata}$, respectively.

We use HuggingFace \texttt{transformers}\footnote{\url{https://huggingface.co/docs/transformers/index}} \citep{wolf2020huggingfaces} for task and language adaptation with PEFT and ICL experiments. For ICL, we employ openICL \citep{wu-etal-2023-openicl} and use $k$-nearest neighbour based retrieval for few-shot demonstrations, following \citet{liu-etal-2022-makes}. For retrieval, we compute sentence level representation of the inference time input and the training data using \citet{reimers-2019-sentence-bert}. We specifically use \texttt{xlm-roberta-base} \citep{conneau-etal-2020-unsupervised} as the base pre-trained model. We choose $8$ input-output pairs as for the few-shot demonstrations. These demonstrations for both tasks are mutually exclusive. For instance, in \textit{Masquerading} with word order, we keep all demonstrations to have re-ordered sentences in the target language. It ensures that the few-shot examples are directly relevant to the task variation with high specificity. 

For PEFT, we utilize  HuggingFace \texttt{PEFT}\footnote{\url{https://github.com/huggingface/peft}} with LoRA \citep{hu2021lora} on top of 4-bit quantization, to fine-tune $\texttt{Llama}_{\texttt{chat}}$ and $\texttt{Airavata}$ on a single 80GB NVIDIA A100 Tensor Core GPU. With \texttt{PEFT-LoRA}, trainable parameters amount to only 0.5\% of the total parameters of the aforementioned LLMs. We train our models with 32-bit paged AdamW \citep{loshchilov2019decoupled} optimizer, with an initial learning rate of $1 \times 10^{-3}$ coupled with a \textit{cosine} scheduler. Refer to Appendix \ref{sec:config_details} for detailed model configuration. 

During inference, we switch to \texttt{Contrastive Search}\footnote{\url{https://huggingface.co/blog/introducing-csearch}} \citep{su2023contrastive} with $\alpha=0.6$ to penalize token repetitions and control model behavior to generate human-level coherent outputs.

\paragraph{Metrics: } We use micro-F1 as our primary evaluation metric for slot filling and named entity recognition, both being NLU tasks. Given that both tasks are framed as text-to-text tasks via an LLM, we also include Exact Match to capture correctness, and chrF++ \citep{chrfplusplus} to assess the lexical overlap between the LLM-generated prediction and the ground-truth reference. Additionally, we measure the naturalness of the generated output on $500$ randomly sampled test instances using MAUVE \citep{pillutla-etal:mauve:neurips2021}.

%% file: results.tex
\section{Results}
In this section, we present our findings with comparative analysis for the approaches of \textit{Handholding, Masquerading, and Bridging} on \texttt{Llama-2} with few-shot ICL and PEFT. For consolidated quantitative figures with PEFT refer to \Cref{table:oraclevspseudo_appen}.

\paragraph{Monolingual ICL Results:} We report near zero performance with \texttt{$\texttt{Llama}_{\texttt{chat}}$} in the monolingual ICL settings. We follow few-shot prompt demonstration under 3 different ICL settings. Here, we provide the input in the target language as is, or \textit{masquerade} it by either transliterating or re-ordering the input. Nevertheless, we observe near-zero micro-F1, exact match (EM) scores, and poor lexical overlap with reference outputs in all three languages for both the tasks. These observations align with the observations made in \citep{razumovskaia2024analyzing} and demonstrate the challenges in adapting a new unseen language in ICL settings to an LLM like \texttt{Llama-2}. 

\begin{table}[h]
\centering
\small
\scalebox{0.9}{
\begin{tabular}{l|cccc}
\toprule
\multirow{2}{*}{\diagbox{Language}{Metric}}& \multicolumn{4}{c}{\texttt{$\texttt{Llama}_{\texttt{chat}}$} \textit{(monolingual)}}\\
\cmidrule(lr){2-5}
& F1  & EM & chrF++ & MAUVE\\
\midrule
\rowcolor{gray!20} \multicolumn{5}{l}{\textbf{Slot Filling}}\\
\midrule
Bengali & $54.72$ & $22.37$ & $71.40$ & $89.07$ \\
Hindi & $51.89$ & $23.15$ & $70.90$ & $59.82$ \\
Tamil & $44.29$ & $14.37$ & $70.65$ & $49.04$ \\
\midrule
\rowcolor{gray!20} \multicolumn{5}{l}{\textbf{Named Entity Recognition}}\\
\midrule
Bengali & $59.98$ & $24.69$ & $85.91$ & $95.28$ \\
Hindi & $71.58$ & $38.25$ & $90.00$ & $98.70$ \\
Tamil & $39.92$ & $12.25$ & $68.72$ & $33.06$ \\
\bottomrule
\end{tabular}
}
\caption{Monolingual performance of $\texttt{Llama}_{\texttt{chat}}$ under PEFT.}
\label{table:monolingual_llama2}
\end{table}

\paragraph{Monolingual PEFT Results:} As shown in \Cref{table:monolingual_llama2}, we observe performance improvements under monolingual settings, when the model parameters are updated with task-specific PEFT. Averaged over both tasks, the exact match (EM) scores of labelled output generations in Bengali, Hindi, and Tamil stand at $23.53\%$, $30.7\%$, and $13.31\%$, respectively. Whereas, the lexical overlap of the generated outputs with the ground-truth outputs are $78.65\%$, $80.45\%$, and $69.68\%$, respectively. These Indic languages are morphologically rich, in general, leading to lower EM scores, though report higher chrF++ (lexical overlap) and MAUVE (naturalness) scores, comparatively.  

\begin{table}[h]
\centering
\small
\scalebox{0.9}{
\begin{tabular}{l|cccc}
\toprule
\multirow{2}{*}{\diagbox{Language}{Metric}}& \multicolumn{4}{c}{\texttt{$\texttt{Llama}_{\texttt{chat}}$} \textit{(Handholding)}}\\
\cmidrule(lr){2-5}
& F1  & EM & chrF++ & MAUVE\\
\midrule
\rowcolor{gray!20} \multicolumn{5}{l}{\textbf{Slot Filling}}\\
\midrule
Bengali & $64.32$ & $36.82$ & $79.27$ & $90.39$ \\
Hindi & $60.60$ & $36.70$ & $77.95$ & $89.72$ \\
Tamil & $61.48$ & $33.79$ & $80.67$ & $76.51$ \\
\midrule
\rowcolor{gray!20} \multicolumn{5}{l}{\textbf{Named Entity Recognition}}\\
\midrule
Bengali & $80.35$ & $45.44$ & $91.00$ & $93.36$ \\
Hindi & $78.03$ & $47.50$ & $90.38$ & $97.09$ \\
Tamil & $74.18$ & $42.69$ & $88.75$ & $81.34$ \\
\bottomrule
\end{tabular}
}
\caption{Effect of \textit{Handholding} on \texttt{$\texttt{Llama}_{\texttt{chat}}$} under PEFT.}
\label{table:handholding_llama2}
\end{table}

\paragraph{Handholding PEFT Results: } \Cref{table:handholding_llama2} shows the performance for the target language under PEFT with \textit{Handholding}. We observe that \textit{Handholding} can help further improve the performance in the target language, with task-specific PEFT. Bengali, Hindi and Tamil benefit from labelled sentence in English under PEFT by $9.6\%$, $8.71\%$, and $17.19\%$ micro-F1 score for slot filling, and $20.37\%$, $6.45\%$, and $34.26\%$ micro-F1 score for named entity recognition. EM scores also improve by an average of $17.6\%$, $11.4\%$, and $24.93\%$ for Bengali. Hindi and Tamil, respectively. Similarly, lexical overlap improves in $6$ out of $6$ cases. However, we observe a drop of $1.92\%$ and $1.61\%$ in naturalness scores of Bengali and Hindi for the NER task.

\begin{table}[h]
    \centering
    \small
    \tabcolsep=0.13cm
    \scalebox{0.9}{
    \begin{tabular}{l|ccc}
    \toprule
    \diagbox{Language}{Change} & {\textit{H}} & $\stackrel{\textit{H + M}}{\textit{(re-ordered)}}$ & $\stackrel{\textit{H + M}}{\textit{(transliterated)}}$\\
    \midrule
    \rowcolor{gray!20} \multicolumn{4}{l}{\textbf{Slot Filling}} \\
    \midrule
    $\textit{en}_{(source)}\rightarrow \textit{bn}_{(target)}$ & $28.02$ & $\underline{30.12^{*}}$ & $18.01$\\
    $\textit{en}_{(source)}\rightarrow \textit{hi}_{(target)}$ & $38.97$ & $\underline{40.82^{*}}$ & $16.57$\\
    $\textit{en}_{(source)}\rightarrow \textit{ta}_{(target)}$ & $22.09$ & $\underline{24.38^{*}}$ & $12.61$\\
    \midrule
    \rowcolor{gray!20} \multicolumn{4}{l}{\textbf{Named Entity Recognition}} \\
    \midrule
    $\textit{en}_{(source)}\rightarrow \textit{bn}_{(target)}$ & $13.89$ &  $\underline{27.88^{*}}$ & $17.78$\\
    $\textit{en}_{(source)}\rightarrow \textit{hi}_{(target)}$ & $47.61$ & $\underline{49.82^{*}}$ & $19.61$\\
    $\textit{en}_{(source)}\rightarrow \textit{ta}_{(target)}$ & $19.07$ & $\underline{30.08^{*}}$ & $18.84$\\
    \bottomrule
    \end{tabular}
    }
    \caption{Micro-F1 scores for the combination of \textit{Handholding (H) and Masquerading (M)} under few-shot ICL.The symbol, $^{*}$ represents statistically significant gains based  on pairwise t-tests with just Handholding ($p < 0.05$).}
    \label{table:icl}
    \end{table} 

\paragraph{Handholding ICL Results: } Similarly, \Cref{table:icl} reports significant improvements in cross-lingual transfer to the target language when using \textit{Handholding} under ICL settings as well. With few-shot ICL using \textit{Handholding}, we see significant gains, as compared to the near-zero performances with few-shot ICL in monolingual settings. Moreover, we are getting non-zero EM scores in $4$ out of 6 cases with \textit{Handholding} under ICL. Nevertheless, as expected, the performance improvements in absolute terms is much higher in \textit{Handholding} with task-specific PEFT (\Cref{table:handholding_llama2}).

\paragraph{Handholding and Masquerading ICL Results:} Further,  \textit{Handholding}, along with \textit{Masquerading} via word re-ordering, leads to statistically significant results under ICL. \Cref{table:icl} shows the results for both \textit{Masquerading} via re-ordering and transliteration. For both the tasks, re-ordering the sentences in all the three languages to resemble the word order in English leads to statistically significant results.  However, \textit{Handholding + Masquerading} via transliterated target sentences under ICL results in performance drops. As shown in \Cref{table:icl}, the  use of transliterated sentences generally results in worse performance than using \textit{Handholding} alone, except for Bengali in NER.


    \begin{table}[h]
    \centering
    \small
    \tabcolsep=0.13cm
    \begin{tabular}{l|cc}
    \toprule
    \diagbox{Language}{Change} & {\textit{H}} & $\stackrel{\textit{H + M}}{\textit{(re-ordered)}}$ \\
    \midrule
    \rowcolor{gray!20} \multicolumn{3}{l}{\textbf{Slot Filling}} \\
    \midrule
    $\textit{en}_{(source)}\rightarrow \textit{bn}_{(target)}$ & $\underline{64.32}$ & $63.19$\\
    $\textit{en}_{(source)}\rightarrow \textit{hi}_{(target)}$ & $60.60$ & $\underline{61.11}$\\
    $\textit{en}_{(source)}\rightarrow \textit{ta}_{(target)}$ & $61.48$ & $\underline{63.30}$\\
    \midrule
    \rowcolor{gray!20} \multicolumn{3}{l}{\textbf{Named Entity Recognition}} \\
    \midrule
    $\textit{en}_{(source)}\rightarrow \textit{bn}_{(target)}$ & $\underline{80.35}$ &  $55.23$\\
    $\textit{en}_{(source)}\rightarrow \textit{hi}_{(target)}$ & $\underline{78.03}$ & $54.01$\\
    $\textit{en}_{(source)}\rightarrow \textit{ta}_{(target)}$ & $\underline{74.18}$ & $43.96$\\
    \bottomrule
    \end{tabular}
    \caption{Micro-F1 scores for the combination of \textit{Handholding (H) and Masquerading (M)} under PEFT.}
    \label{table:hnmpeft}
    \end{table}

\paragraph{Handholding and Masquerading PEFT Results:} As shown in \Cref{table:handholding_llama2} and \Cref{table:icl}, \textit{Handholding} benefits the target language, both under ICL and PEFT settings. Similarly, combining \textit{Handholding} with \textit{Masquerading} via word re-ordering has shown to be beneficial under ICL. \Cref{table:hnmpeft} presents the results for the combination of \textit{Handholding} and \textit{Masquerading} with task-specific PEFT. However, the benefits from \textit{Masquerading} appear to diminish or be counterproductive during PEFT, especially for NER tasks. Nevertheless we see statistically significant gains for Slot Filling in Tamil, though not for Hindi. Within \textit{Masquerading}, we do not explore the setting of transliteration of target sentence due to its consistent poor performance under few-shot ICL. For slot filling, Bengali sees a reduction of $1.13\%$ micro-F1 whereas Hindi and Tamil observe increase in micro-F1 scores by $0.51\%$ and $1.82\%$, respectively.

    \begin{table}[h]
    \centering
    \small
    \begin{tabular}{l|cc}
    \toprule
    \diagbox{Language}{Model} & {\texttt{$\texttt{Llama}_{\texttt{chat}}$}} & {\texttt{Airavata}}\\
    \midrule
    \rowcolor{gray!20} \multicolumn{3}{l}{\textbf{Slot Filling}} \\
    \midrule
    $\textit{bn}_{(target)}$ & $54.72$ & $\underline{64.28}^{*}$ \\
    $\textit{ta}_{(target)}$ & $44.29$ & $\underline{46.03}^{*}$ \\
    \midrule
    \rowcolor{gray!20} \multicolumn{3}{l}{\textbf{Named Entity Recognition}} \\
    \midrule
    $\textit{bn}_{(target)}$ & $59.98$ &  $\underline{66.62}^{*}$ \\
    $\textit{ta}_{(target)}$ & $39.92$ & $\underline{66.14}^{*}$ \\
    \bottomrule
    \end{tabular}
    \caption{Micro-F1 scores for the effect of \textit{Bridging} on monolingual performance in Bengali and Tamil. The symbol, $^{*}$ represents statistically significant gains for \texttt{Airavata} based  on pairwise t-tests with  \texttt{$\texttt{Llama}_{\texttt{chat}}$} ($p < 0.05$).}

    \label{table:b_finetune}
    \end{table}

\paragraph{Bridging: }In \textit{Bridging}, Hindi serves as the bridge language, while English still remains the predominant language. In this case, we evaluate model performance on Bengali and Tamil as the target languages. As discussed in \Cref{sec:imple_details}, we use \texttt{Airavata} to evaluate the effect of increased representation of Hindi on the related languages of Bengali and Tamil. Our first observation follows that \textit{Bridging} improves monolingual performance in both Bengali and Tamil with task-specific PEFT. As shown in \Cref{table:b_finetune}, \texttt{Airavata} outperforms \texttt{$\texttt{Llama}_{\texttt{chat}}$} in both Bengali and Tamil for both tasks of slot filling and named entity recognition. For slot filling, Bengali observes an increase of $9.56\%$ micro-F1, $21.37\%$ increase in EM score, $10.17\%$ increase in lexical overlap and an improved output naturalness by $9.63\%$. Whereas, Tamil benefits with an increased micro-F1, and EM of $1.74\%$, and $7.03\%$. respectively. However, lexical overlap and naturalness of generated outputs with reference outputs falls by $9.31\%$ and $12.52\%$ in Airavata as compared to \texttt{$\texttt{Llama}_{\texttt{chat}}$}. For named entity recognition, we see similar improvements under all metrics, for both languages post \textit{Bridging} except the fall in naturalness for Bengali by $2.47\%$.

    \begin{table}[h]
    \centering
    \small
    \begin{tabular}{l|cc}
    \toprule
    \diagbox{Language}{Model} & {\texttt{$\texttt{Llama}_{\texttt{chat}}$}} & {\texttt{Airavata}}\\
    \midrule
    \rowcolor{gray!20} \multicolumn{3}{l}{\textbf{Slot Filling}} \\
    \midrule
    $\textit{en}_{(source)}\rightarrow \textit{bn}_{(target)}$ & $64.32$ & $\underline{67.21}$ \\
    $\textit{en}_{(source)}\rightarrow \textit{ta}_{(target)}$ & $61.48$ & $\underline{65.24}$ \\
    \midrule
    \rowcolor{gray!20} \multicolumn{3}{l}{\textbf{Named Entity Recognition}} \\
    \midrule
    $\textit{en}_{(source)}\rightarrow \textit{bn}_{(target)}$ & $80.35$ &  $\underline{84.80}$ \\
    $\textit{en}_{(source)}\rightarrow \textit{ta}_{(target)}$ & $74.18$ & $\underline{82.09}$ \\
    \bottomrule
    \end{tabular}
    \caption{Micro-F1 scores for the combination of \textit{Handholding (H) + Bridging (B)} under PEFT.}
    
    \label{table:finetune}
    \end{table}

\paragraph{Handholding and Bridging: }\Cref{table:finetune} presents the best performing combination, in terms of model performance for slot filling and named entity recognition. This is achieved by \textit{Bridging} \texttt{Llama-2} with Hindi, followed by task-specific model adaptation through PEFT with \textit{Handholding}. In this case, Bengali benefits by $2.89\%$ micro-F1, $11.72\%$ EM score, $1.54\%$ lexical overlap and $4.98\%$ in naturalness as compared to \textit{Handholding} with \texttt{$\texttt{Llama}_{\texttt{chat}}$} for the task of slot-filling and $4.45\%$ in micro-F1, $13.81\%$ in EM score, $2.86\%$ in lexical overlap and $6.49\%$ in naturalness for named entity recognition. Similarly, for slot filling, Tamil observes increase of $3.84\%$ micro-F1, $10.37\%$ EM score, but a drop in $0.26\%$ lexical overlap and $2.69\%$ naturalness of generated output. Whereas, for named entity recognition, model performance in Tamil increases by $7.91\%$ micro-F1, $19.87\%$ EM score, $5.89\%$ lexical overlap, and $18.12\%$ naturalness score.

%% file: conclusion.tex
\section{Conclusion}

In this work, through extensive experiments on English-centric \texttt{Llama-2-7b-chat} under both ICL and PEFT, we show that \textit{Handholding} improves NLU and NLG in low-resource languages: Bengali, Hindi and Tamil by exploiting cross-lingual transfer from English, demonstrating that the predominant language of an LLM can be leveraged to aid low-resource languages. Further, \textit{Bridging} with a low-resource related language Hindi, results to improved monolingual task performance in related languages of Bengali and Tamil. Ultimately, through \textit{Handholding + Bridging}, we show that incorporating both the predominant language of the LLM and adapting the LLM in a related language results to better cross-lingual transfer, leading to significantly improved understanding and generation in other related low-resource languages. However, adapting the target language to resemble the predominant language in terms of syntax and script \textit{(Masquerading)}, only leads to superficial performance improvements in the low-resource language. 

%% file: limitations.tex
\section*{Limitations}
The very notion of the cross-lingual transfer objective from an labelled sentence in source language to an unannotated sentence in target language requires parallel data. High-quality parallel data is not uniformly available for all language pairs, specifically for underrepresented language families like the Indic family. The requirement of an annotated source during training and/or inference adds up as a bottleneck. As shown in \Cref{sec:imple_details}, it can be subdued if we have a reference model to label the source, before cross-lingual transfer. However, the likelihood of a high-accuracy reference model is minimal when considering the case of cross-lingual transfer of annotations between two underrepresented languages.

%% file: appendix.tex
\appendix

\section{Evaluation Results}
Refer to \Cref{table:oraclevspseudo_appen} for micro-F1, EM and lexical overlap scores for all experiments with \textit{Handholding}, \textit{Masquerading} and \textit{Bridging} under PEFT. 

\begin{table*}[h]
\centering
\tiny
\scalebox{0.65}{
\begin{tabular}{l|cccccccccccccccccccc}
\toprule
\multirow{5}{*}{\diagbox{Language}{Configuration}}& \multicolumn{12}{c}{\texttt{Llama-2}} & \multicolumn{8}{c}{\texttt{Airavata}}\\
\cmidrule(lr){2-13}
\cmidrule(lr){14-21}
& \multicolumn{4}{c}{\textit{monolingual}} & \multicolumn{4}{c}{\textit{H}} & \multicolumn{4}{c}{\textit{H + M}} & \multicolumn{4}{c}{\textit{B (monolingual)}} & \multicolumn{4}{c}{\textit{H + B}}\\
\cmidrule(lr){2-5}
\cmidrule(lr){6-9}
\cmidrule(lr){10-13}
\cmidrule(lr){14-17}
\cmidrule(lr){18-21}
& F1  & EM & chrF++ & MAUVE & F1  & EM & chrF++ & MAUVE & F1  & EM & chrF++ & MAUVE & F1  & EM & chrF++  & MAUVE & F1  & EM & chrF++ & MAUVE\\
\midrule
\rowcolor{gray!20} \multicolumn{21}{l}{\textit{Slot Filling}}\\
\midrule
 Bengali & $54.72$ & $22.37$ & $71.40$ & $89.07$ & $64.32$ & $36.82$ &$79.27$ & $90.39$ & $63.19$ & $0.96$ & $71.81$ & $37.6$ & $64.28$ & $43.74$ & $\underline{81.57}$ & $\underline{98.70}$ & $\underline{67.21}$ & $\underline{48.54}$ & $80.81$ & $95.37$\\
 Hindi & $51.89$ &$23.15$ & $70.90$  & $59.82$ & $60.60$ & $\underline{36.70}$ & $\underline{77.95}$  & $\underline{89.72}$& $\underline{61.11}$ & $17.29$ & $73.49$ & $24.18$ & $-$ & $-$ & $-$ & $-$& $-$ & $-$ & $-$ & $-$\\
 Tamil & $44.29$& $14.37$& $70.65$ & $49.04$ & $61.48$ & $33.79$ & $\underline{80.67}$ & $\underline{76.51}$& $63.30$ & $17.80$ & $74.96$ & $19.67$ & $46.03$ & $21.40$ & $61.34$ & $36.52$& $\underline{65.24}$ & $\underline{44.16}$ & $80.41$ & $73.82$\\
\midrule
\rowcolor{gray!20} \multicolumn{21}{l}{\textit{Named Entity Recognition}}\\
\midrule
 Bengali  & $59.98$ & $24.69$ & $85.91$ & $95.28$ & $80.35$ & $45.44$ & $91.00$ & $93.36$& $55.23$ & $0.37$ & $54.43$ & $15.14$& $66.42$ & $34.63$ & $89.45$ & $92.81$& $\underline{84.80}$ & $\underline{59.25}$ & $\underline{93.86}$ & $\underline{99.85}$\\
 Hindi & $71.58$ & $38.25$ & $90.00$ & $\underline{98.70}$ & $\underline{78.03}$ & $\underline{47.50}$ & $\underline{90.38}$ & $97.09$& $54.01$ & $0.63$ & $46.18$ & $18.62$& $-$ & $-$ & $-$ & & $-$ & $-$ & $-$ & $-$\\
 Tamil  & $39.92$ & $12.25$ & $68.72$ & $33.06$ & $74.18$ & $42.69$ & $88.75$ & $81.34$& $43.96$ & $1.31$ & $49.93$ & $45.28$& $66.14$ & $42.81$ & $91.42$ & $99.22$& $\underline{82.09}$ & $\underline{62.56}$ & $\underline{94.64}$& $\underline{99.46}$\\
\bottomrule
\end{tabular}
}
\caption{micro-F1, EM, chrF++, and MAUVE scores under PEFT with the model configurations of \textit{H: Handholding}, \textit{M: Masquerading}, and \textit{B: Bridging}. Here, MAUVE is computed on $500$ randomly sampled test instances.}
\label{table:oraclevspseudo_appen}
\end{table*}

\section{Dataset Splits}
The dataset split for both tasks is presented in \Cref{table:dataset_split}. For Massive, we use the train, validation, and test split as on HuggingFace \texttt{datasets}\footnote{\url{https://huggingface.co/datasets/MASSIVE}}. For evaluation, we restrict the test set to only contain utterances that have at least $1$ token with a slot label. For Naamapadam, we split the $16$k sampled instances in a $8$:$1$:$1$ ratio to create train, validation, and test subsets.

\begin{table}[h]
    \centering
    \small
    \begin{tabular}{l|cc}
    \toprule
    \diagbox{Task}{Dataset Split} & {\texttt{Train}} & {\texttt{Test}}\\
    \midrule
    Slot Filling & $11.5\text{k}$ & $1.9\text{k}$ \\
    \midrule
    Named Entity Recognition & $12.8\text{k}$ & $1.6\text{k}$ \\
    \bottomrule
    \end{tabular}
    \caption{Dataset split for slot filling and named entity recognition tasks.}
    \label{table:dataset_split}
\end{table}

\section{List of Label Types}
Complete list of label types within Massive and Naamapadam is showcased in \Cref{table:slot-types}.

\section{Training and Inference Configuration}
\label{sec:config_details}
We present our PEFT and ICL hyperparameter settings in Table \ref{train_config}. These hyperparameters remain the same across both \texttt{Llama-2-7b-chat} and \texttt{Airavata-7b}.

\section{Prompt Details}
Refer to \Cref{train_prompt1,train_prompt2,train_prompt_incontext} for prompts used in our experiments.

\begin{table}[h]
\centering
\tiny
\tabcolsep=0.06cm
\begin{tabular}{lll}
    \toprule
    \texttt{date} & \texttt{time} & \texttt{color\_type} \\
    \texttt{house\_place} & \texttt{place\_name} & \texttt{time\_zone} \\
    \texttt{artist\_name} & \texttt{timeofday} & \texttt{meal\_type} \\
    \texttt{food\_type} & \texttt{order\_type} & \texttt{news\_topic} \\
    \texttt{music\_genre} & \texttt{weather\_descriptor} & \texttt{playlist\_name} \\
    \texttt{device\_type} & \texttt{player\_setting} & \texttt{song\_name} \\
    \texttt{media\_type} & \texttt{joke\_type} & \texttt{alarm\_type} \\
    \texttt{music\_descriptor} & \texttt{business\_name} & \texttt{business\_type} \\
    \texttt{general\_frequency} & \texttt{change\_amount} & \texttt{event\_name} \\
    \texttt{ingredient} & \texttt{person} & \texttt{coffee\_type} \\
    \texttt{drink\_type} & \texttt{music\_album} & \texttt{relation} \\
    \texttt{radio\_name} & \texttt{app\_name} & \texttt{podcast\_descriptor} \\
    \texttt{audiobook\_author} & \texttt{audiobook\_name} & \texttt{cooking\_type} \\
    \texttt{list\_name} & \texttt{game\_name} & \texttt{podcast\_name} \\
    \texttt{movie\_type} & \texttt{movie\_name} & \texttt{transport\_type} \\
    \texttt{transport\_name} & \texttt{transport\_agency} & \texttt{transport\_descriptor} \\
    \texttt{definition\_word} & \texttt{currency\_name} & \texttt{personal\_info} \\
    \texttt{email\_address} & \texttt{email\_folder} & \texttt{game\_type} \\
    \texttt{change\_amount} & & \\
    \midrule
    \texttt{person (PER)} & \texttt{organization (ORG)} & \texttt{location (LOC)} \\
    \bottomrule
\end{tabular}
\caption{List of all label types in Massive and Naamapadam, in that order.}
\label{table:slot-types}
\end{table}

\begin{table}[h]
\centering
\tiny
\tabcolsep=0.06cm
\begin{tabular}{lll}
\toprule
 & \textit{Massive} & \textit{Naamapadam} \\
\midrule
\texttt{LoRA rank} & \texttt{8} & \texttt{8}\\
\texttt{LoRA alpha} & \texttt{16} & \texttt{16} \\
\texttt{Batch size (Training)} & \texttt{32} & \texttt{16} \\
\texttt{Batch size (Inference)} & \texttt{4} & \texttt{4} \\
\texttt{Gradient checkpointing} & \texttt{True} & \texttt{True} \\
\texttt{Gradient accumulation steps} & \texttt{4} & \texttt{4} \\
\texttt{Max. gradient norm} & \texttt{0.3} & \texttt{0.3} \\
\texttt{Epochs} & \texttt{2, 3} & \texttt{3} \\
\texttt{Learning rate} & \texttt{1e-3} & \texttt{1e-3} \\
\texttt{Optimizer} & \texttt{32-bit AdamW (paged)} & \texttt{32-bit Adam (paged)}\\
\texttt{Precision} & \texttt{bf16} & \texttt{bf16}\\
\texttt{LR scheduler} & \texttt{cosine} & \texttt{cosine} \\
\texttt{Train batch size} & \texttt{32} & \texttt{16} \\
\texttt{Warm-up ratio} & \texttt{0.05} & \texttt{0.05} \\
\texttt{Max. sequence length (Training)} & \texttt{512} & \texttt{1024}\\
\texttt{Stopping Criteria (Inference)} & \texttt{512} & \texttt{768} \\
\texttt{Penalty alpha (Inference)} & \texttt{0.6} & \texttt{0.6} \\
\texttt{top\_k (Inference)} & \texttt{4} & \texttt{4}\\
\bottomrule
\end{tabular}
\label{train_config}
\caption{Complete set of hyperparameters for PEFT and ICL. For ICL, we use the same inference-time hyperparameters as mentioned above.}
\label{train_config}
\end{table}

\begin{table}[h]
\centering
\begin{tabular}{p{7cm}}
\toprule
\texttt{Reinsert the slot annotations into the following Hindi sentence using the information in the English sentence.} \\
\\
\noindent \texttt{\#\#\# Hindi: [Unannotated target]} \\
\noindent \texttt{\#\#\# English: [Annotated source]} \\
\noindent \texttt{\#\#\# Output:} \\
\bottomrule
\end{tabular}
\caption{Example prompt format for PEFT with the cross-lingual annotation transfer objective.}
\label{train_prompt1}
\end{table}

\begin{table}[h]
\centering
\begin{tabular}{p{7cm}}
\toprule
\texttt{Reinsert the slot annotations into the following Hindi sentence.} \\
\\
\noindent \texttt{\#\#\# Hindi: [Unannotated target]} \\
\noindent \texttt{\#\#\# Output:} \\
\bottomrule
\end{tabular}
\caption{Prompt format for PEFT with the monolingual annotation objective.}
\label{train_prompt2}
\end{table}

\begin{table}
\centering
\begin{tabular}{p{7cm}}
\toprule
\texttt{<<SYS>> Add annotations for the corresponding tokens in Tamil sentences using the annotation information given in the English sentence. The annotations are marked in the format [annotation\_type : token/value]}\\
\texttt{Input will be provided in the following format}\\
\texttt{\#\#\# Tamil: Tamil sentence}\\
\texttt{\#\#\# English: English sentence}\\
\texttt{Output should be printed after the string ``\#\#\# Output:"}\\
\texttt{The final output should be the Tamil sentence with annotations inserted corresponding to the annotations of the English sentence. Do not add any extra annotations to the Tamil sentence, which are not present in the English sentence input.<</SYS>>}\\
\\
\texttt{Add annotations for the given tokens \textit{<list of tokens present in annotated source>} in Tamil sentence using the annotation information given in the English sentence}\\
\texttt{\#\#\# Tamil: \textit{[Unannotated target]}}   \\
\texttt{\#\#\# English: \textit{[Annotated source]}}\\
\texttt{\#\#\# Output: \textit{[Annotated target]}} \\
.\\
.\\
.\\
\textit{$\times$ n few-shot examples}\\
\\
\texttt{Add annotations for the given tokens \textit{<list of tokens present in annotated source>} in Tamil sentence using the annotation information given in the English sentence}\\
\texttt{\#\#\# Tamil: \textit{<An unannotated Tamil sentence>}}   \\
\texttt{\#\#\# English: \textit{<An annotated English sentence>}}\\
\texttt{\#\#\# Output:}\\
\bottomrule
\end{tabular}
\caption{Example prompt format for few-shot ICL with the cross-lingual annotation transfer objective.}
\label{train_prompt_incontext}
\end{table}

%% file: emnlp2023.bbl
\begin{thebibliography}{35}
\expandafter\ifx\csname natexlab\endcsname\relax\def\natexlab#1{#1}\fi

\bibitem[{Awasthi et~al.(2023)Awasthi, Gupta, Samanta, Dave, Sarawagi, and Talukdar}]{awasthi2023bootstrapping}
Abhijeet Awasthi, Nitish Gupta, Bidisha Samanta, Shachi Dave, Sunita Sarawagi, and Partha Talukdar. 2023.
\newblock \href {http://arxiv.org/abs/2210.07313} {Bootstrapping multilingual semantic parsers using large language models}.

\bibitem[{Brown et~al.(1993)Brown, Della~Pietra, Della~Pietra, and Mercer}]{brown-etal-1993-mathematics}
Peter~F. Brown, Stephen~A. Della~Pietra, Vincent~J. Della~Pietra, and Robert~L. Mercer. 1993.
\newblock \href {https://aclanthology.org/J93-2003} {The mathematics of statistical machine translation: Parameter estimation}.
\newblock \emph{Computational Linguistics}, 19(2):263--311.

\bibitem[{Brown et~al.(2020)Brown, Mann, Ryder, Subbiah, Kaplan, Dhariwal, Neelakantan, Shyam, Sastry, Askell, Agarwal, Herbert-Voss, Krueger, Henighan, Child, Ramesh, Ziegler, Wu, Winter, Hesse, Chen, Sigler, Litwin, Gray, Chess, Clark, Berner, McCandlish, Radford, Sutskever, and Amodei}]{brown2020language}
Tom~B. Brown, Benjamin Mann, Nick Ryder, Melanie Subbiah, Jared Kaplan, Prafulla Dhariwal, Arvind Neelakantan, Pranav Shyam, Girish Sastry, Amanda Askell, Sandhini Agarwal, Ariel Herbert-Voss, Gretchen Krueger, Tom Henighan, Rewon Child, Aditya Ramesh, Daniel~M. Ziegler, Jeffrey Wu, Clemens Winter, Christopher Hesse, Mark Chen, Eric Sigler, Mateusz Litwin, Scott Gray, Benjamin Chess, Jack Clark, Christopher Berner, Sam McCandlish, Alec Radford, Ilya Sutskever, and Dario Amodei. 2020.
\newblock \href {http://arxiv.org/abs/2005.14165} {Language models are few-shot learners}.

\bibitem[{Chowdhery et~al.(2022)Chowdhery, Narang, Devlin, Bosma, Mishra, Roberts, Barham, Chung, Sutton, Gehrmann, Schuh, Shi, Tsvyashchenko, Maynez, Rao, Barnes, Tay, Shazeer, Prabhakaran, Reif, Du, Hutchinson, Pope, Bradbury, Austin, Isard, Gur-Ari, Yin, Duke, Levskaya, Ghemawat, Dev, Michalewski, Garcia, Misra, Robinson, Fedus, Zhou, Ippolito, Luan, Lim, Zoph, Spiridonov, Sepassi, Dohan, Agrawal, Omernick, Dai, Pillai, Pellat, Lewkowycz, Moreira, Child, Polozov, Lee, Zhou, Wang, Saeta, Diaz, Firat, Catasta, Wei, Meier-Hellstern, Eck, Dean, Petrov, and Fiedel}]{chowdhery2022palm}
Aakanksha Chowdhery, Sharan Narang, Jacob Devlin, Maarten Bosma, Gaurav Mishra, Adam Roberts, Paul Barham, Hyung~Won Chung, Charles Sutton, Sebastian Gehrmann, Parker Schuh, Kensen Shi, Sasha Tsvyashchenko, Joshua Maynez, Abhishek Rao, Parker Barnes, Yi~Tay, Noam Shazeer, Vinodkumar Prabhakaran, Emily Reif, Nan Du, Ben Hutchinson, Reiner Pope, James Bradbury, Jacob Austin, Michael Isard, Guy Gur-Ari, Pengcheng Yin, Toju Duke, Anselm Levskaya, Sanjay Ghemawat, Sunipa Dev, Henryk Michalewski, Xavier Garcia, Vedant Misra, Kevin Robinson, Liam Fedus, Denny Zhou, Daphne Ippolito, David Luan, Hyeontaek Lim, Barret Zoph, Alexander Spiridonov, Ryan Sepassi, David Dohan, Shivani Agrawal, Mark Omernick, Andrew~M. Dai, Thanumalayan~Sankaranarayana Pillai, Marie Pellat, Aitor Lewkowycz, Erica Moreira, Rewon Child, Oleksandr Polozov, Katherine Lee, Zongwei Zhou, Xuezhi Wang, Brennan Saeta, Mark Diaz, Orhan Firat, Michele Catasta, Jason Wei, Kathy Meier-Hellstern, Douglas Eck, Jeff Dean, Slav Petrov, and Noah Fiedel. 2022.
\newblock \href {http://arxiv.org/abs/2204.02311} {Palm: Scaling language modeling with pathways}.

\bibitem[{Chung et~al.(2022)Chung, Hou, Longpre, Zoph, Tay, Fedus, Li, Wang, Dehghani, Brahma, Webson, Gu, Dai, Suzgun, Chen, Chowdhery, Castro-Ros, Pellat, Robinson, Valter, Narang, Mishra, Yu, Zhao, Huang, Dai, Yu, Petrov, Chi, Dean, Devlin, Roberts, Zhou, Le, and Wei}]{chung2022scaling}
Hyung~Won Chung, Le~Hou, Shayne Longpre, Barret Zoph, Yi~Tay, William Fedus, Yunxuan Li, Xuezhi Wang, Mostafa Dehghani, Siddhartha Brahma, Albert Webson, Shixiang~Shane Gu, Zhuyun Dai, Mirac Suzgun, Xinyun Chen, Aakanksha Chowdhery, Alex Castro-Ros, Marie Pellat, Kevin Robinson, Dasha Valter, Sharan Narang, Gaurav Mishra, Adams Yu, Vincent Zhao, Yanping Huang, Andrew Dai, Hongkun Yu, Slav Petrov, Ed~H. Chi, Jeff Dean, Jacob Devlin, Adam Roberts, Denny Zhou, Quoc~V. Le, and Jason Wei. 2022.
\newblock \href {http://arxiv.org/abs/2210.11416} {Scaling instruction-finetuned language models}.

\bibitem[{Conneau et~al.(2020)Conneau, Khandelwal, Goyal, Chaudhary, Wenzek, Guzm{\'a}n, Grave, Ott, Zettlemoyer, and Stoyanov}]{conneau-etal-2020-unsupervised}
Alexis Conneau, Kartikay Khandelwal, Naman Goyal, Vishrav Chaudhary, Guillaume Wenzek, Francisco Guzm{\'a}n, Edouard Grave, Myle Ott, Luke Zettlemoyer, and Veselin Stoyanov. 2020.
\newblock \href {https://doi.org/10.18653/v1/2020.acl-main.747} {Unsupervised cross-lingual representation learning at scale}.
\newblock In \emph{Proceedings of the 58th Annual Meeting of the Association for Computational Linguistics}, pages 8440--8451, Online. Association for Computational Linguistics.

\bibitem[{Cui et~al.(2024)Cui, Yang, and Yao}]{cui2024efficient}
Yiming Cui, Ziqing Yang, and Xin Yao. 2024.
\newblock \href {http://arxiv.org/abs/2304.08177} {Efficient and effective text encoding for chinese llama and alpaca}.

\bibitem[{FitzGerald et~al.(2022)FitzGerald, Hench, Peris, Mackie, Rottmann, Sanchez, Nash, Urbach, Kakarala, Singh, Ranganath, Crist, Britan, Leeuwis, Tur, and Natarajan}]{fitzgerald2022massive}
Jack FitzGerald, Christopher Hench, Charith Peris, Scott Mackie, Kay Rottmann, Ana Sanchez, Aaron Nash, Liam Urbach, Vishesh Kakarala, Richa Singh, Swetha Ranganath, Laurie Crist, Misha Britan, Wouter Leeuwis, Gokhan Tur, and Prem Natarajan. 2022.
\newblock \href {http://arxiv.org/abs/2204.08582} {Massive: A 1m-example multilingual natural language understanding dataset with 51 typologically-diverse languages}.

\bibitem[{Gala et~al.(2024)Gala, Jayakumar, Husain, M, Khan, Kanojia, Puduppully, Khapra, Dabre, Murthy, and Kunchukuttan}]{gala2024airavata}
Jay Gala, Thanmay Jayakumar, Jaavid~Aktar Husain, Aswanth~Kumar M, Mohammed Safi Ur~Rahman Khan, Diptesh Kanojia, Ratish Puduppully, Mitesh~M. Khapra, Raj Dabre, Rudra Murthy, and Anoop Kunchukuttan. 2024.
\newblock Airavata: Introducing hindi instruction-tuned llm.
\newblock \emph{arXiv preprint arXiv: 2401.15006}.

\bibitem[{Gupta et~al.(2023)Gupta, Thérien, Ibrahim, Richter, Anthony, Belilovsky, Rish, and Lesort}]{gupta2023continual}
Kshitij Gupta, Benjamin Thérien, Adam Ibrahim, Mats~L. Richter, Quentin Anthony, Eugene Belilovsky, Irina Rish, and Timothée Lesort. 2023.
\newblock \href {http://arxiv.org/abs/2308.04014} {Continual pre-training of large language models: How to (re)warm your model?}

\bibitem[{Hu et~al.(2021)Hu, Shen, Wallis, Allen-Zhu, Li, Wang, Wang, and Chen}]{hu2021lora}
Edward~J. Hu, Yelong Shen, Phillip Wallis, Zeyuan Allen-Zhu, Yuanzhi Li, Shean Wang, Lu~Wang, and Weizhu Chen. 2021.
\newblock \href {http://arxiv.org/abs/2106.09685} {Lora: Low-rank adaptation of large language models}.

\bibitem[{Husain et~al.(2024)Husain, Dabre, Kumar, Gala, Jayakumar, Puduppully, and Kunchukuttan}]{husain2024romansetu}
Jaavid~Aktar Husain, Raj Dabre, Aswanth Kumar, Jay Gala, Thanmay Jayakumar, Ratish Puduppully, and Anoop Kunchukuttan. 2024.
\newblock \href {http://arxiv.org/abs/2401.14280} {Romansetu: Efficiently unlocking multilingual capabilities of large language models models via romanization}.

\bibitem[{Joshi et~al.(2020)Joshi, Santy, Budhiraja, Bali, and Choudhury}]{joshi-etal-2020-state}
Pratik Joshi, Sebastin Santy, Amar Budhiraja, Kalika Bali, and Monojit Choudhury. 2020.
\newblock \href {https://doi.org/10.18653/v1/2020.acl-main.560} {The state and fate of linguistic diversity and inclusion in the {NLP} world}.
\newblock In \emph{Proceedings of the 58th Annual Meeting of the Association for Computational Linguistics}, pages 6282--6293, Online. Association for Computational Linguistics.

\bibitem[{Kubis et~al.(2023)Kubis, Sk{\'o}rzewski, Sowa{\'n}ski, and Zi{\k{e}}tkiewicz}]{kubis2023back}
Marek Kubis, Pawe{\l} Sk{\'o}rzewski, Marcin Sowa{\'n}ski, and Tomasz Zi{\k{e}}tkiewicz. 2023.
\newblock \href {http://arxiv.org/abs/2310.16609} {Back transcription as a method for evaluating robustness of natural language understanding models to speech recognition errors}.
\newblock \emph{arXiv preprint arXiv:2310.16609}.

\bibitem[{Li et~al.(2023)Li, Koto, Wu, Aji, and Baldwin}]{li2023bactrianx}
Haonan Li, Fajri Koto, Minghao Wu, Alham~Fikri Aji, and Timothy Baldwin. 2023.
\newblock \href {http://arxiv.org/abs/2305.15011} {Bactrian-x: Multilingual replicable instruction-following models with low-rank adaptation}.

\bibitem[{Liu et~al.(2022)Liu, Shen, Zhang, Dolan, Carin, and Chen}]{liu-etal-2022-makes}
Jiachang Liu, Dinghan Shen, Yizhe Zhang, Bill Dolan, Lawrence Carin, and Weizhu Chen. 2022.
\newblock \href {https://doi.org/10.18653/v1/2022.deelio-1.10} {What makes good in-context examples for {GPT}-3?}
\newblock In \emph{Proceedings of Deep Learning Inside Out (DeeLIO 2022): The 3rd Workshop on Knowledge Extraction and Integration for Deep Learning Architectures}, pages 100--114, Dublin, Ireland and Online. Association for Computational Linguistics.

\bibitem[{Loshchilov and Hutter(2019)}]{loshchilov2019decoupled}
Ilya Loshchilov and Frank Hutter. 2019.
\newblock \href {http://arxiv.org/abs/1711.05101} {Decoupled weight decay regularization}.

\bibitem[{Mesnard et~al.(2024)Mesnard, Hardin, Dadashi, Bhupatiraju, Pathak, Sifre, Rivière, Kale, Love, Tafti, Hussenot, Sessa, Chowdhery, Roberts, Barua, Botev, Castro-Ros, Slone, Héliou, Tacchetti, Bulanova, Paterson, Tsai, Shahriari, Lan, Choquette-Choo, Crepy, Cer, Ippolito, Reid, Buchatskaya, Ni, Noland, Yan, Tucker, Muraru, Rozhdestvenskiy, Michalewski, Tenney, Grishchenko, Austin, Keeling, Labanowski, Lespiau, Stanway, Brennan, Chen, Ferret, Chiu, Mao-Jones, Lee, Yu, Millican, Sjoesund, Lee, Dixon, Reid, Mikuła, Wirth, Sharman, Chinaev, Thain, Bachem, Chang, Wahltinez, Bailey, Michel, Yotov, Chaabouni, Comanescu, Jana, Anil, McIlroy, Liu, Mullins, Smith, Borgeaud, Girgin, Douglas, Pandya, Shakeri, De, Klimenko, Hennigan, Feinberg, Stokowiec, hui Chen, Ahmed, Gong, Warkentin, Peran, Giang, Farabet, Vinyals, Dean, Kavukcuoglu, Hassabis, Ghahramani, Eck, Barral, Pereira, Collins, Joulin, Fiedel, Senter, Andreev, and Kenealy}]{gemmateam2024gemma}
Thomas Mesnard, Cassidy Hardin, Robert Dadashi, Surya Bhupatiraju, Shreya Pathak, Laurent Sifre, Morgane Rivière, Mihir~Sanjay Kale, Juliette Love, Pouya Tafti, Léonard Hussenot, Pier~Giuseppe Sessa, Aakanksha Chowdhery, Adam Roberts, Aditya Barua, Alex Botev, Alex Castro-Ros, Ambrose Slone, Amélie Héliou, Andrea Tacchetti, Anna Bulanova, Antonia Paterson, Beth Tsai, Bobak Shahriari, Charline~Le Lan, Christopher~A. Choquette-Choo, Clément Crepy, Daniel Cer, Daphne Ippolito, David Reid, Elena Buchatskaya, Eric Ni, Eric Noland, Geng Yan, George Tucker, George-Christian Muraru, Grigory Rozhdestvenskiy, Henryk Michalewski, Ian Tenney, Ivan Grishchenko, Jacob Austin, James Keeling, Jane Labanowski, Jean-Baptiste Lespiau, Jeff Stanway, Jenny Brennan, Jeremy Chen, Johan Ferret, Justin Chiu, Justin Mao-Jones, Katherine Lee, Kathy Yu, Katie Millican, Lars~Lowe Sjoesund, Lisa Lee, Lucas Dixon, Machel Reid, Maciej Mikuła, Mateo Wirth, Michael Sharman, Nikolai Chinaev, Nithum Thain, Olivier Bachem, Oscar Chang,
  Oscar Wahltinez, Paige Bailey, Paul Michel, Petko Yotov, Rahma Chaabouni, Ramona Comanescu, Reena Jana, Rohan Anil, Ross McIlroy, Ruibo Liu, Ryan Mullins, Samuel~L Smith, Sebastian Borgeaud, Sertan Girgin, Sholto Douglas, Shree Pandya, Siamak Shakeri, Soham De, Ted Klimenko, Tom Hennigan, Vlad Feinberg, Wojciech Stokowiec, Yu~hui Chen, Zafarali Ahmed, Zhitao Gong, Tris Warkentin, Ludovic Peran, Minh Giang, Clément Farabet, Oriol Vinyals, Jeff Dean, Koray Kavukcuoglu, Demis Hassabis, Zoubin Ghahramani, Douglas Eck, Joelle Barral, Fernando Pereira, Eli Collins, Armand Joulin, Noah Fiedel, Evan Senter, Alek Andreev, and Kathleen Kenealy. 2024.
\newblock \href {http://arxiv.org/abs/2403.08295} {Gemma: Open models based on gemini research and technology}.

\bibitem[{Mhaske et~al.(2023)Mhaske, Kedia, Doddapaneni, Khapra, Kumar, Murthy, and Kunchukuttan}]{mhaske-etal-2023-naamapadam}
Arnav Mhaske, Harshit Kedia, Sumanth Doddapaneni, Mitesh~M. Khapra, Pratyush Kumar, Rudra Murthy, and Anoop Kunchukuttan. 2023.
\newblock \href {https://doi.org/10.18653/v1/2023.acl-long.582} {Naamapadam: A large-scale named entity annotated data for {I}ndic languages}.
\newblock In \emph{Proceedings of the 61st Annual Meeting of the Association for Computational Linguistics (Volume 1: Long Papers)}, pages 10441--10456, Toronto, Canada. Association for Computational Linguistics.

\bibitem[{Och and Ney(2003)}]{och-ney-2003-systematic}
Franz~Josef Och and Hermann Ney. 2003.
\newblock \href {https://doi.org/10.1162/089120103321337421} {A systematic comparison of various statistical alignment models}.
\newblock \emph{Computational Linguistics}, 29(1):19--51.

\bibitem[{Ojo et~al.(2024)Ojo, Ogueji, Stenetorp, and Adelani}]{ojo2024good}
Jessica Ojo, Kelechi Ogueji, Pontus Stenetorp, and David~Ifeoluwa Adelani. 2024.
\newblock \href {http://arxiv.org/abs/2311.07978} {How good are large language models on african languages?}

\bibitem[{Ouyang et~al.(2022)Ouyang, Wu, Jiang, Almeida, Wainwright, Mishkin, Zhang, Agarwal, Slama, Ray, Schulman, Hilton, Kelton, Miller, Simens, Askell, Welinder, Christiano, Leike, and Lowe}]{ouyang2022training}
Long Ouyang, Jeff Wu, Xu~Jiang, Diogo Almeida, Carroll~L. Wainwright, Pamela Mishkin, Chong Zhang, Sandhini Agarwal, Katarina Slama, Alex Ray, John Schulman, Jacob Hilton, Fraser Kelton, Luke Miller, Maddie Simens, Amanda Askell, Peter Welinder, Paul Christiano, Jan Leike, and Ryan Lowe. 2022.
\newblock \href {http://arxiv.org/abs/2203.02155} {Training language models to follow instructions with human feedback}.

\bibitem[{Pillutla et~al.(2021)Pillutla, Swayamdipta, Zellers, Thickstun, Welleck, Choi, and Harchaoui}]{pillutla-etal:mauve:neurips2021}
Krishna Pillutla, Swabha Swayamdipta, Rowan Zellers, John Thickstun, Sean Welleck, Yejin Choi, and Zaid Harchaoui. 2021.
\newblock Mauve: Measuring the gap between neural text and human text using divergence frontiers.
\newblock In \emph{NeurIPS}.

\bibitem[{Popovi{\'c}(2017)}]{chrfplusplus}
Maja Popovi{\'c}. 2017.
\newblock \href {https://doi.org/10.18653/v1/W17-4770} {chr{F}++: words helping character n-grams}.
\newblock In \emph{Proceedings of the Second Conference on Machine Translation}, pages 612--618, Copenhagen, Denmark. Association for Computational Linguistics.

\bibitem[{Ramesh et~al.(2022)Ramesh, Doddapaneni, Bheemaraj, Jobanputra, AK, Sharma, Sahoo, Diddee, J, Kakwani, Kumar, Pradeep, Nagaraj, Deepak, Raghavan, Kunchukuttan, Kumar, and Khapra}]{ramesh-etal-2022-samanantar}
Gowtham Ramesh, Sumanth Doddapaneni, Aravinth Bheemaraj, Mayank Jobanputra, Raghavan AK, Ajitesh Sharma, Sujit Sahoo, Harshita Diddee, Mahalakshmi J, Divyanshu Kakwani, Navneet Kumar, Aswin Pradeep, Srihari Nagaraj, Kumar Deepak, Vivek Raghavan, Anoop Kunchukuttan, Pratyush Kumar, and Mitesh~Shantadevi Khapra. 2022.
\newblock \href {https://doi.org/10.1162/tacl_a_00452} {Samanantar: The largest publicly available parallel corpora collection for 11 {I}ndic languages}.
\newblock \emph{Transactions of the Association for Computational Linguistics}, 10:145--162.

\bibitem[{Rathore et~al.(2023)Rathore, Dhingra, Singla, and {Mausam}}]{rathore-etal-2023-zgul}
Vipul Rathore, Rajdeep Dhingra, Parag Singla, and {Mausam}. 2023.
\newblock \href {https://doi.org/10.18653/v1/2023.emnlp-main.431} {{ZGUL}: Zero-shot generalization to unseen languages using multi-source ensembling of language adapters}.
\newblock In \emph{Proceedings of the 2023 Conference on Empirical Methods in Natural Language Processing}, pages 6969--6987, Singapore. Association for Computational Linguistics.

\bibitem[{Razumovskaia et~al.(2024)Razumovskaia, Vulić, and Korhonen}]{razumovskaia2024analyzing}
Evgeniia Razumovskaia, Ivan Vulić, and Anna Korhonen. 2024.
\newblock \href {http://arxiv.org/abs/2403.01929} {Analyzing and adapting large language models for few-shot multilingual nlu: Are we there yet?}

\bibitem[{Reimers and Gurevych(2019)}]{reimers-2019-sentence-bert}
Nils Reimers and Iryna Gurevych. 2019.
\newblock \href {https://arxiv.org/abs/1908.10084} {Sentence-bert: Sentence embeddings using siamese bert-networks}.
\newblock In \emph{Proceedings of the 2019 Conference on Empirical Methods in Natural Language Processing}. Association for Computational Linguistics.

\bibitem[{Sabet et~al.(2021)Sabet, Dufter, Yvon, and Schütze}]{sabet2021simalign}
Masoud~Jalili Sabet, Philipp Dufter, François Yvon, and Hinrich Schütze. 2021.
\newblock \href {http://arxiv.org/abs/2004.08728} {Simalign: High quality word alignments without parallel training data using static and contextualized embeddings}.

\bibitem[{Su and Collier(2023)}]{su2023contrastive}
Yixuan Su and Nigel Collier. 2023.
\newblock \href {http://arxiv.org/abs/2210.14140} {Contrastive search is what you need for neural text generation}.

\bibitem[{Tanwar et~al.(2023)Tanwar, Dutta, Borthakur, and Chakraborty}]{tanwar2023multilingual}
Eshaan Tanwar, Subhabrata Dutta, Manish Borthakur, and Tanmoy Chakraborty. 2023.
\newblock \href {http://arxiv.org/abs/2305.05940} {Multilingual llms are better cross-lingual in-context learners with alignment}.

\bibitem[{Touvron et~al.(2023)Touvron, Martin, Stone, Albert, Almahairi, Babaei, Bashlykov, Batra, Bhargava, Bhosale, Bikel, Blecher, Ferrer, Chen, Cucurull, Esiobu, Fernandes, Fu, Fu, Fuller, Gao, Goswami, Goyal, Hartshorn, Hosseini, Hou, Inan, Kardas, Kerkez, Khabsa, Kloumann, Korenev, Koura, Lachaux, Lavril, Lee, Liskovich, Lu, Mao, Martinet, Mihaylov, Mishra, Molybog, Nie, Poulton, Reizenstein, Rungta, Saladi, Schelten, Silva, Smith, Subramanian, Tan, Tang, Taylor, Williams, Kuan, Xu, Yan, Zarov, Zhang, Fan, Kambadur, Narang, Rodriguez, Stojnic, Edunov, and Scialom}]{touvron2023llama}
Hugo Touvron, Louis Martin, Kevin Stone, Peter Albert, Amjad Almahairi, Yasmine Babaei, Nikolay Bashlykov, Soumya Batra, Prajjwal Bhargava, Shruti Bhosale, Dan Bikel, Lukas Blecher, Cristian~Canton Ferrer, Moya Chen, Guillem Cucurull, David Esiobu, Jude Fernandes, Jeremy Fu, Wenyin Fu, Brian Fuller, Cynthia Gao, Vedanuj Goswami, Naman Goyal, Anthony Hartshorn, Saghar Hosseini, Rui Hou, Hakan Inan, Marcin Kardas, Viktor Kerkez, Madian Khabsa, Isabel Kloumann, Artem Korenev, Punit~Singh Koura, Marie-Anne Lachaux, Thibaut Lavril, Jenya Lee, Diana Liskovich, Yinghai Lu, Yuning Mao, Xavier Martinet, Todor Mihaylov, Pushkar Mishra, Igor Molybog, Yixin Nie, Andrew Poulton, Jeremy Reizenstein, Rashi Rungta, Kalyan Saladi, Alan Schelten, Ruan Silva, Eric~Michael Smith, Ranjan Subramanian, Xiaoqing~Ellen Tan, Binh Tang, Ross Taylor, Adina Williams, Jian~Xiang Kuan, Puxin Xu, Zheng Yan, Iliyan Zarov, Yuchen Zhang, Angela Fan, Melanie Kambadur, Sharan Narang, Aurelien Rodriguez, Robert Stojnic, Sergey Edunov, and Thomas
  Scialom. 2023.
\newblock \href {http://arxiv.org/abs/2307.09288} {Llama 2: Open foundation and fine-tuned chat models}.

\bibitem[{Wolf et~al.(2020)Wolf, Debut, Sanh, Chaumond, Delangue, Moi, Cistac, Rault, Louf, Funtowicz, Davison, Shleifer, von Platen, Ma, Jernite, Plu, Xu, Scao, Gugger, Drame, Lhoest, and Rush}]{wolf2020huggingfaces}
Thomas Wolf, Lysandre Debut, Victor Sanh, Julien Chaumond, Clement Delangue, Anthony Moi, Pierric Cistac, Tim Rault, Rémi Louf, Morgan Funtowicz, Joe Davison, Sam Shleifer, Patrick von Platen, Clara Ma, Yacine Jernite, Julien Plu, Canwen Xu, Teven~Le Scao, Sylvain Gugger, Mariama Drame, Quentin Lhoest, and Alexander~M. Rush. 2020.
\newblock \href {http://arxiv.org/abs/1910.03771} {Huggingface's transformers: State-of-the-art natural language processing}.

\bibitem[{Wu et~al.(2023)Wu, Wang, Ye, Wu, Feng, Xu, and Qiao}]{wu-etal-2023-openicl}
Zhenyu Wu, Yaoxiang Wang, Jiacheng Ye, Zhiyong Wu, Jiangtao Feng, Jingjing Xu, and Yu~Qiao. 2023.
\newblock \href {https://doi.org/10.18653/v1/2023.acl-demo.47} {{O}pen{ICL}: An open-source framework for in-context learning}.
\newblock In \emph{Proceedings of the 61st Annual Meeting of the Association for Computational Linguistics (Volume 3: System Demonstrations)}, pages 489--498, Toronto, Canada. Association for Computational Linguistics.

\bibitem[{Zhao et~al.(2024)Zhao, Zhang, Gao, Zhang, Gui, and Huang}]{zhao2024llama}
Jun Zhao, Zhihao Zhang, Luhui Gao, Qi~Zhang, Tao Gui, and Xuanjing Huang. 2024.
\newblock \href {http://arxiv.org/abs/2401.01055} {Llama beyond english: An empirical study on language capability transfer}.

\end{thebibliography}
